\documentclass[conference]{IEEEtran}
\IEEEoverridecommandlockouts
\usepackage{cite}
\usepackage{amsmath,amssymb,amsfonts}
\usepackage{algorithmic}
\usepackage{graphicx}
\usepackage{textcomp}
\usepackage{xcolor}
\usepackage{booktabs}
\usepackage{multirow} 
\usepackage{subcaption}
\def\BibTeX{{\rm B\kern-.05em{\sc i\kern-.025em b}\kern-.08em
    T\kern-.1667em\lower.7ex\hbox{E}\kern-.125emX}}
\begin{document}

\title{Global-Local Aware Scene Text Editing}

\author{
\IEEEauthorblockN{Fuxiang Yang\textsuperscript{1}, Tonghua Su\textsuperscript{1$^{*}$}, Donglin Di\textsuperscript{2}, Yin Chen\textsuperscript{1}, Xiangqian Wu\textsuperscript{1,3}, Zhongjie Wang\textsuperscript{1}, and Lei Fan\textsuperscript{4$^{*}$}}\thanks{$^{*}$ Corresponding authors.}
\IEEEauthorblockA{
\textsuperscript{1}Harbin Institute of Technology, Harbin, China\\
\textsuperscript{2}Li Auto, Beijing, China\\
\textsuperscript{3}Suzhou Research Institute, HIT, Suzhou, China \\
\textsuperscript{4}University of New South Wales, Sydney, Australia
}
\IEEEauthorblockA{
hityangfx@foxmail.com, \{thsu, chenyin, 
xqwu, rainy\}@hit.edu.cn, donglin.ddl@gmail.com, lei.fan1@unsw.edu.au}
}

\maketitle

\begin{abstract}
Scene Text Editing (STE) involves replacing text in a scene image with new target text while preserving both the original text style and background texture.
Existing methods suffer from two major challenges:  inconsistency and length-insensitivity.
They often fail to maintain coherence between the edited local patch and the surrounding area, and they struggle to handle significant differences in text length before and after editing.
To tackle these challenges, we propose an end-to-end framework called Global-Local Aware Scene Text Editing (GLASTE), which simultaneously incorporates high-level global contextual information along with delicate local features.
Specifically, we design a global-local combination structure, joint global and local losses, and enhance text image features to ensure consistency in text style within local patches while maintaining harmony between local and global areas.
Additionally, we express the text style as a vector independent of the image size, which can be transferred to target text images of various sizes.
We use an affine fusion to fill target text images into the editing patch while maintaining their aspect ratio unchanged.
Extensive experiments on real-world datasets validate that our GLASTE model outperforms previous methods in both quantitative metrics and qualitative results and effectively mitigates the two challenges.
\end{abstract}


\section{Introduction}
\label{sec:intro}

Scene Text Editing (STE) aims to re-render the target text in the style of a specified text patch within a given scene image, producing a natural and realistic replacement for that text patch~\cite{wu2019editing}.
STE has broad application potential in areas such as image translation~\cite{vistransICPR2024} and creative text design~\cite{chen2023textdiffuser,ji2023improving}. Traditionally, STE tasks were performed manually by designers using various image processing tools, making the process labor-intensive and costly.
With the advancement of deep learning \cite{fan2025grainbrain,fan2023identifying,fan2024patch} and generative models \cite{wang2024towards,sun2024eggen,fan2022fast}, leveraging these models has emerged as a promising and feasible solution for automated STE.

\begin{figure}[t]
    \centering
    \includegraphics[width=0.48\textwidth]{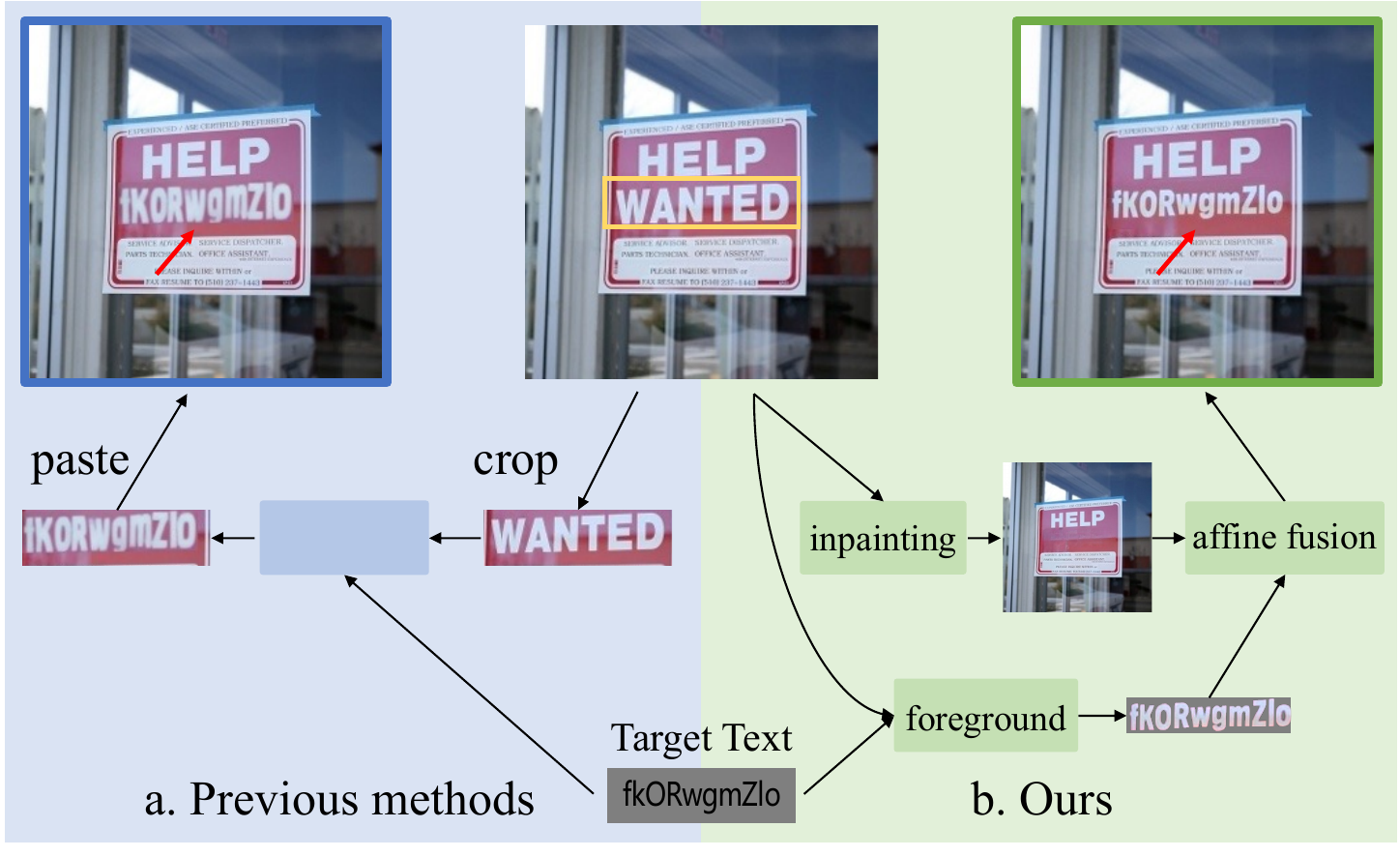}
    \caption{
     (a) shows the running process of the previous STE methods, where the generation model processes text patches, which relies on the ``crop-and-paste'' operation.
    (b) Our GLASTE method directly uses the entire image as input, inpainting the specified text region and then rendering the target text within that area to generate a scene image.
    }
    \label{fig:intro}
\end{figure}

Previous works~\cite{wu2019editing,qu2023exploring,krishnan2023textstylebrush,CLASTE,ji2023improving} employed GANs and diffusion models to automate STE. For example, SRNet~\cite{wu2019editing} decomposes the task into content, style, and background components, using synthetic paired text samples for training. TextStyleBrush\cite{krishnan2023textstylebrush} extended this approach to real-world scene data through self-supervised training, using the input image as supervision and eliminating the need for synthetic paired datasets. MOSTEL~\cite{qu2023exploring} further combines synthetic and real-world data. These methods rely on a ``crop-and-paste'' strategy. The target region is cropped from the image, content in the target region is modified using generative models, and the edited content is reintegrated into the original image, as shown in Figure~\ref{fig:intro}.

There are two primary challenges in these methods: inconsistency and length-insensitivity. 
Specifically, \textbf{inconsistency} refers to the lack of coherence between the edited text patch and its surrounding area, causing visual disharmony. 
For example, while the generated text may appear realistic at the local patch level, integrating it back into the original image can produce visual artifacts such as darker or blurrier edges around the edited regions.
\textbf{Length-insensitivity} refers to the difficulty in text length differences between the source and target content. 
When the target text is shorter or longer, the edited text may appear stretched or overcrowded, distorting the character structure~\cite{krishnan2023textstylebrush}. The first challenge arises because the target text patch is cropped and modified without considering global context, particularly the surrounding pixels, resulting in discrepancies in overall harmony, local structure, and background consistency. The second challenge stems from the absence of explicit size control mechanisms in existing methods, where the model relies solely on data-driven learning to map the source text to the target text, making it difficult to handle variations in text length effectively.

In this paper, we adopt a global-local joint perspective. Unlike the traditional crop-and-paste strategy, we first perform inpainting on the target region to generate a coherent background image, then use the entire image along with the target content as input to generate the modified text. The inpainted background is combined with the modified content to produce the final result. Additionally, we introduce a size-independent strategy that encodes text style as a size-independent vector, allowing the editing patch to retain a preserved margin instead of stretching the text to fill the entire region. An affine transformation is applied to explicitly control the rendering size of the target text, ensuring more natural and visually consistent typesetting.

We propose a framework named Global-Local Aware Scene Text Editing (GLASTE), consisting of three modules: the inpainting module, the foreground module, and the affine fusion module. 
Specifically, the inpainting module uses a Fourier-based operation~\cite{suvorov2022resolution} to treat the edited local patch as the area to be inpainted. It recovers the edited regions to generate seamless areas, which are further refined and integrated into the background inpainted image.
The foreground module consists of a style encoder, a content encoder, and a text synthesizer. It takes the scene image as input and extracts style features from the edited patch using the style encoder. Simultaneously, the target content is processed by the content encoder to obtain content features. The text synthesizer then combines these style and content features to generate a foreground-styled text image.
The affine fusion module performs an affine transformation on the foreground-styled text image and blends it with the background inpainted image to produce the final edited result.

Our contributions are summarized as follows:
\begin{itemize}
\item  
We propose an end-to-end framework, GLASTE, which incorporates a global-local structure to generate consistent and readable text while ensuring visual harmony in the edited areas.

\item 
Our model adaptively generates target text images of varying lengths, effectively avoiding deformation caused by stretching or squeezing text in the edited area.

\item 
Extensive experiments conducted on real-world scene data demonstrate that GLASTE outperforms previous methods both qualitatively and quantitatively.

\end{itemize}

\begin{figure*}[t]
    \centering
    \includegraphics[width=0.88\textwidth]{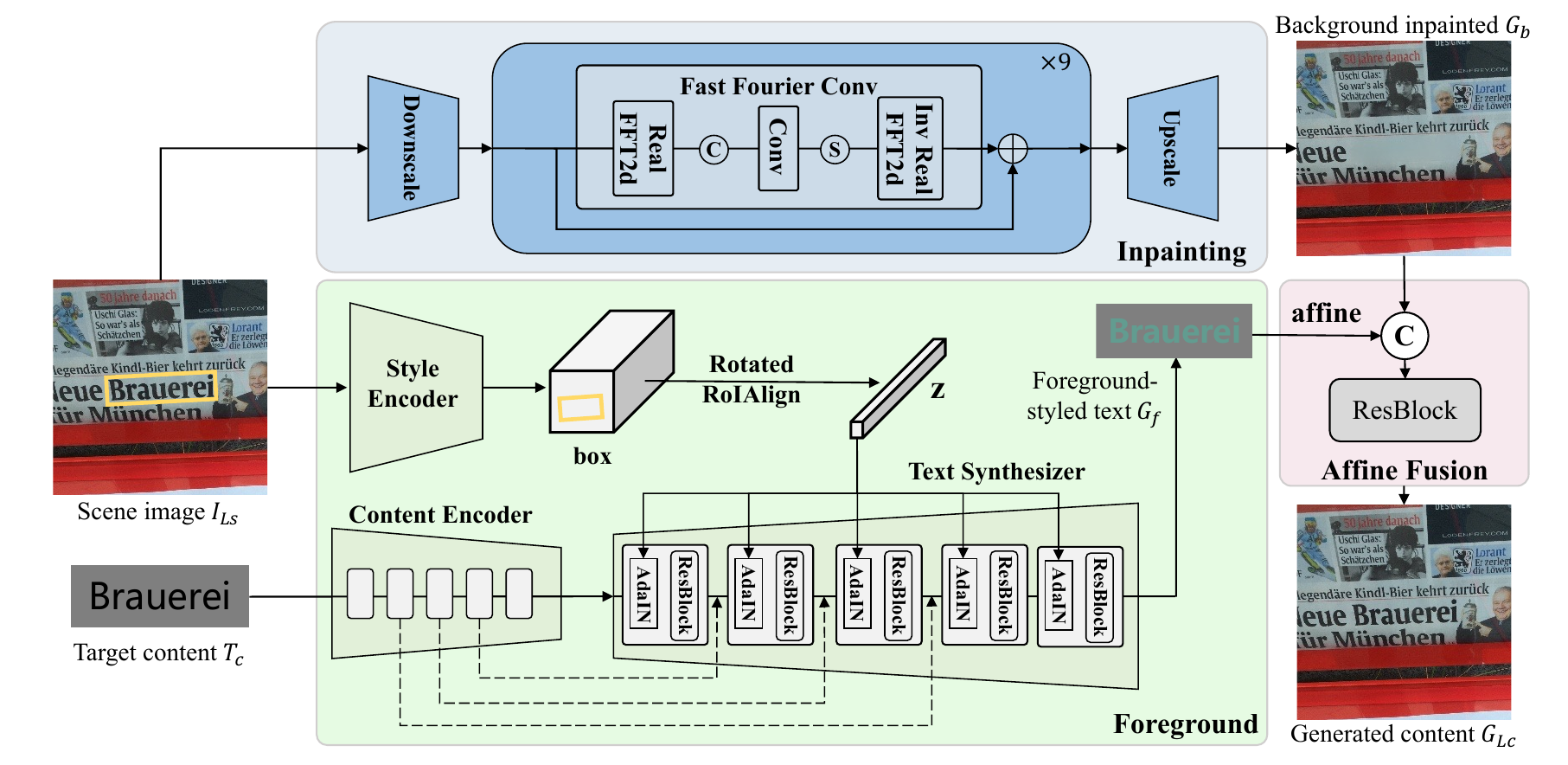}
    \caption{\label{fig:architecture}  The overall structure of GLASTE. 
    The network consists of an inpainting module, a foreground module, and an affine fusion module.
    The foreground module includes a style encoder, a content encoder, and a text synthesizer.
    }
    \vspace{-1.2em}
    \end{figure*}

\section{Method}

Our Global-Local Aware Scene Text Editing model (GLASTE) consists of three modules: inpainting, foreground, and affine fusion, as shown in Figure~\ref{fig:architecture}. Given a scene image \( I_{Ls} \in \mathbb{R}^{H_L \times W_L \times 3} \) with a corresponding target region \( box = (c_{xb}, c_{yb}, w_b, h_b, \theta_b) \), \( c_{xb} \) and \( c_{yb} \) represent the center point, $(H_L, W_L)$  and (\( w_b \), \( h_b \)) denote the width and height of the scene image and target region box, and \(\theta_b\) is the rotation angle. Our goal is to modify the source content of \( box \) in \( I_{Ls} \) using the target content \( T_c \in \mathbb{R}^{H_c \times W_c \times 3} \).  
Specifically, the inpainting module masks the box region in the scene image \( I_{Ls} \) and recover the masked region to generate background inpainted image \( G_b \in \mathbb{R}^{H_L \times W_L \times 3} \) through an inpainting process. The foreground module extracts the style vector \( \mathbf{z} \in \mathbb{R}^{512} \) from the box region in \( I_{Ls} \) using a style encoder and Rotated RoIAlign operation, while content features are extracted from \( T_c \) using a content encoder. The text synthesizer combines the style vector \( \mathbf{z} \) and content features to generate the foreground-styled text \( G_f \in \mathbb{R}^{H_f \times W_f \times 3} \). Finally, the background inpainted image \( G_b \) and the foreground-styled text \( G_f \) are merged using affine fusion to produce the final image \( G_{Lc} \), maintaining the same size as the input scene image \( I_{Ls} \).

\subsection{Inpainting Module}

 The inpainting module is designed to fill the target region in \( I_{Ls} \) and generate a clean background texture \( G_b \). It leverages Fast Fourier Convolution (FFC)~\cite{NEURIPS2020_2fd5d41e,suvorov2022resolution} to provide a receptive field that covers the entire image. 
The module takes \( I_{Ls} \) as input and applies downsampling to produce a feature map \( \mathbf{FB} \in \mathbb{R}^{H \times W \times C} ,H=\frac{H_L}{8},W=\frac{W_L}{8}\). This feature map is then transformed into the frequency domain as follows:
\begin{equation}
x = C(y_r, y_i) = C(\phi( \mathbf{FB} )),
\end{equation}
where \( x \in \mathbb{R}^{H \times (\frac{W}{2} + 1) \times 2C} \), \( x_r \) and \( x_i \) represent the real and imaginary parts, \( C \) denotes the concatenation operation, and \( \phi \) is the Real FFT2d function. The transformed \( x \) is processed by a convolutional block to produce \( x' \in \mathbb{R}^{H \times (\frac{W}{2} + 1) \times 2C} \).
Subsequently, \( x' \) is transformed back to the spatial domain:
\begin{equation}
y = \phi^{-1}(x'_r, x'_i) = \phi^{-1}(C^{-1}(x')),
\end{equation}
where \( y \in \mathbb{R}^{H \times W \times C} \), \( C^{-1} \) denotes the split operation that separates \( x' \) into its real part \( x'_r \) and imaginary part \( x'_i \), and \( \phi^{-1} \) represents the Inverse Real FFT2d operation. 
Multiple FFCs are connected in a residual manner to further extract features. The features are then upsampled to produce the background inpainted image \( G_b \).
The frequency-domain representation allows the module to infer missing pixels in the target region effectively by utilizing global pixel information. Furthermore, since the text of the target region is removed from the input \( I_{Ls} \), the method avoids residual shadows of the original text.

\subsection{Foreground Module}
The foreground module comprises a style encoder, a content encoder, and a text synthesizer.
It extracts the style from the target region in the style scene image $I_{Ls}$ and transfers it to the target text image $T_{c}$, resulting in the foreground-styled text image $G_{f}$.
Specifically, The style encoder employs ResNet34 to extract a feature map \(\mathbf{FS} \in \mathbb{R}^{\frac{H_L}{16} \times \frac{W_L}{16} \times 512}\) from the input images \(I_{Ls}\), effectively capturing the style features of the target regions. To accommodate text regions with varying rotation angles, Rotated RoIAlign is applied to \(\mathbf{FS}\) using a predefined bounding box. 
This process is formulated as follows:
\begin{equation}
    \mathbf{z} = \pi(\psi(\mathbf{FS}, \tau(box_{scaled}))),
\end{equation}
where \(box_{scaled}\) represents the predefined bounding box, scaled according to the resolution of the feature map. The function \(\tau(\cdot)\) maps the scaled bounding box into a rotated grid of sampling points within the feature map space. The function \(\psi( \cdot)\) performs bilinear interpolation on \(\mathbf{FS}\) at the specified sampling points to extract feature values. Finally, \(\pi(\cdot)\) aggregates the sampled features through average pooling to produce a fixed-size output \(\mathbf{z} \in \mathbb{R}^{ 512}\), which encodes the text style features of the rotated target region.
 This process eliminates cropping and discards patch width and height information, ensuring the style vector \(\mathbf{z}\) is independent of the target region size.  
 
The content encoder employs  ResNet34, which takes the target content image \(T_c\) as input and generates feature maps \(\mathbf{FC}_i^d \in \mathbb{R}^{\frac{H_c}{2^i} \times \frac{W_c}{2^i} \times C_i}\), where \(i = 1, 2, 3, 4, 5\) corresponds to the \(i\)-th convolutional stage.  
The text synthesizer, consisting of 5 residual blocks, employs a structure opposite to that of the content encoder. It progressively upsamples the input content feature map \(\mathbf{FC}\) while integrating text style information \(\mathbf{z}\) through AdaIN~\cite{huang2017arbitrary}, ultimately generating the styled foreground text \(G_f\). AdaIN is defined as:  
\begin{equation}
    \text{AdaIN}(\mathbf{FC}^u_j, \mathbf{z}) = \mathbf{z}_s \left( \frac{\mathbf{FC}^u_j - \mu(\mathbf{FC}^u_j)}{\sigma(\mathbf{FC}^u_j)} \right) + \mathbf{z}_b,
\end{equation}
where \(\mathbf{z}_s\) and \(\mathbf{z}_b\) are the scale and bias factors derived from \(\mathbf{z}\) through a linear layer, and \(\mu\) and \(\sigma\) represent the mean and standard deviation, respectively. 
The feature map after the \(j\)-th residual block is represented as \(\mathbf{FC}^u_j \in \mathbb{R}^{(W_c \times 2^{j-5}) \times (H_c \times 2^{j-5}) \times C_j}\), where \(j = 1, 2, 3, 4, 5\).  
To enhance text image features, skip connections are introduced. Feature maps generated during the content encoder's downsampling are merged with those of corresponding resolutions in the text synthesizer during upsampling, expressed as:  
$\mathbf{FC}^u_k = \text{C}(\mathbf{FC}^u_k, \mathbf{FC}^d_{5-k}), \quad k = 1, 2, 3$.
It is beneficial for improving recognition accuracy in the generated results.

\subsection{Affine Fusion Module}
The affine fusion module combines the foreground-styled text \( G_f \) with the background inpainted image \( G_b \) to generate \( G_c \). The merging process involves an affine transformation that adjusts \( G_f \) to match the dimensions of \( G_b \). This transformation, parameterized by \( \theta \) in the normalized coordinate system, is computed as:  
\begin{equation}
    \theta = \left(T_2 \cdot M^{-1} \cdot T_1^{-1}\right)^{-1},
\end{equation}
where \( T_1 \) and \( T_2 \) are the normalization matrices for the original and target coordinates, respectively, and the affine matrix \( M \) is determined based on $box$.  
The module then employs several residual blocks and concludes with a final layer, outputting $G_c$.
By applying different \( T_1 \) and \( T_2 \), the affine transformation reduces length insensitivity.
In long-to-short text transitions, \( G_f \) is scaled to match the height of the target region, while in short-to-long transitions, \( G_f \) is scaled to match the width of the target region.

\subsection{Joint Global and Local Losses}

In order to model global and local aspects and improve the overall quality of the model generation, we construct global and local loss functions separately. 
The global loss function refers to $G_{Lc}$ and uses the input scene image $I_{Ls}$ as ground truth. 
The local loss function, on the other hand, extracts local text patch images $I_{s} \in \mathbb{R} ^{h\times w\times 3}$ and $G_{c}\in \mathbb{R} ^{h\times w\times 3}$ and $G_{c}$  using Rotated RoIAlign from $I_{Ls}$ and $G_{Lc}$, using $I_{s}$ as the ground truth for $G_{c}$.

We use two discriminators $\mathbf{D}_{1}$ and $\mathbf{D}_{2}$ to respectively discern the authenticity of the generated global images and local patches, and set weight $\alpha$ to balance the discriminator losses. 
Each discriminator is a PatchGAN \cite{isola2017image}, consisting of four convolutional blocks with strides of 2, and one convolutional block with a stride of 1.
The discriminator loss can be denoted as follows:
\begin{align}
  \mathcal{L}_{D} &= E\left(\log \mathbf{D}_{1}\left(I_{Ls}\right)
  +\log \left(1-\mathbf{D}_{1}\left(G_{Lc}\right)\right)\right) \notag \\
  &+ \alpha E\left(\log \mathbf{D}_{2}\left(I_{s}\right)
  +  \log \left(1-\mathbf{D}_{2}\left(G_{c}\right)\right)\right).
\end{align}

In addition, we evaluate the quality of generating large images using the L1 loss $\mathcal{L}_{1}$, perceptual loss $\mathcal{L}_{Per}$, text recognition loss $\mathcal{L}_{Rec}$.
The $\mathcal{L}_{1}$ and $\mathcal{L}_{Per}$ consider both global and local regions simultaneously, and we multiply the local loss by weight $\beta$, thereby emphasizing the generation of the text regions.

The $\mathcal{L}_{1}$ loss can be expressed as follows:
\begin{equation}
  \mathcal{L}_{1}=   \| {G}_{Lc} - \mathcal{I}_{Ls} \|_{1}
+ \beta \| {G}_{c} - \mathcal{I}_{s} \|_{1}.
\end{equation}

The $\mathcal{L}_{Per}$ \cite{johnson2016perceptual} uses a pre-trained VGG19 \cite{simonyan2014very} model, measuring the distance between the output feature maps $\phi(\cdot)$ of specific layers:
\begin{align}
  \mathcal{L}_{Per}&=\sum_{i} \left\|\phi_{i}\left(G_{Lc}\right)-\phi_{i}\left(I_{Ls}\right)\right\|_{1} \notag \\
&+ \beta \sum_{i} \left\|\phi_{i}\left(G_{c}\right)-\phi_{i}\left(I_{s}\right)\right\|_{1},
\end{align}

The $\mathcal{L}_{Rec}$ uses a text recognition model $\mathbf{R}$ based on CRNN \cite{shi2016end} to measure the CTC loss between two strings:
\begin{equation}
    \begin{split}
    \mathcal{L}_{Rec} & =\sum_{i} (\text { CTC }\left(\mathbf{R}(G_{c_{i}}), S_{c_{i}}\right)).
    \end{split}
\end{equation}

The overall loss of the model is:
\begin{equation}
  {\ell}_{total}= \mathcal{L}_{D}  + 
  \lambda_{1}\mathcal{L}_{1}+\lambda_{2}\mathcal{L}_{Per} + \lambda_{3}\mathcal{L}_{Rec}.
\end{equation}
where $\lambda_{1}, \lambda_ {2}, \lambda_{3}$ are hyper-parameters balancing the losses.

\begin{figure}[t]
    \centering
\includegraphics[width=0.45\textwidth]{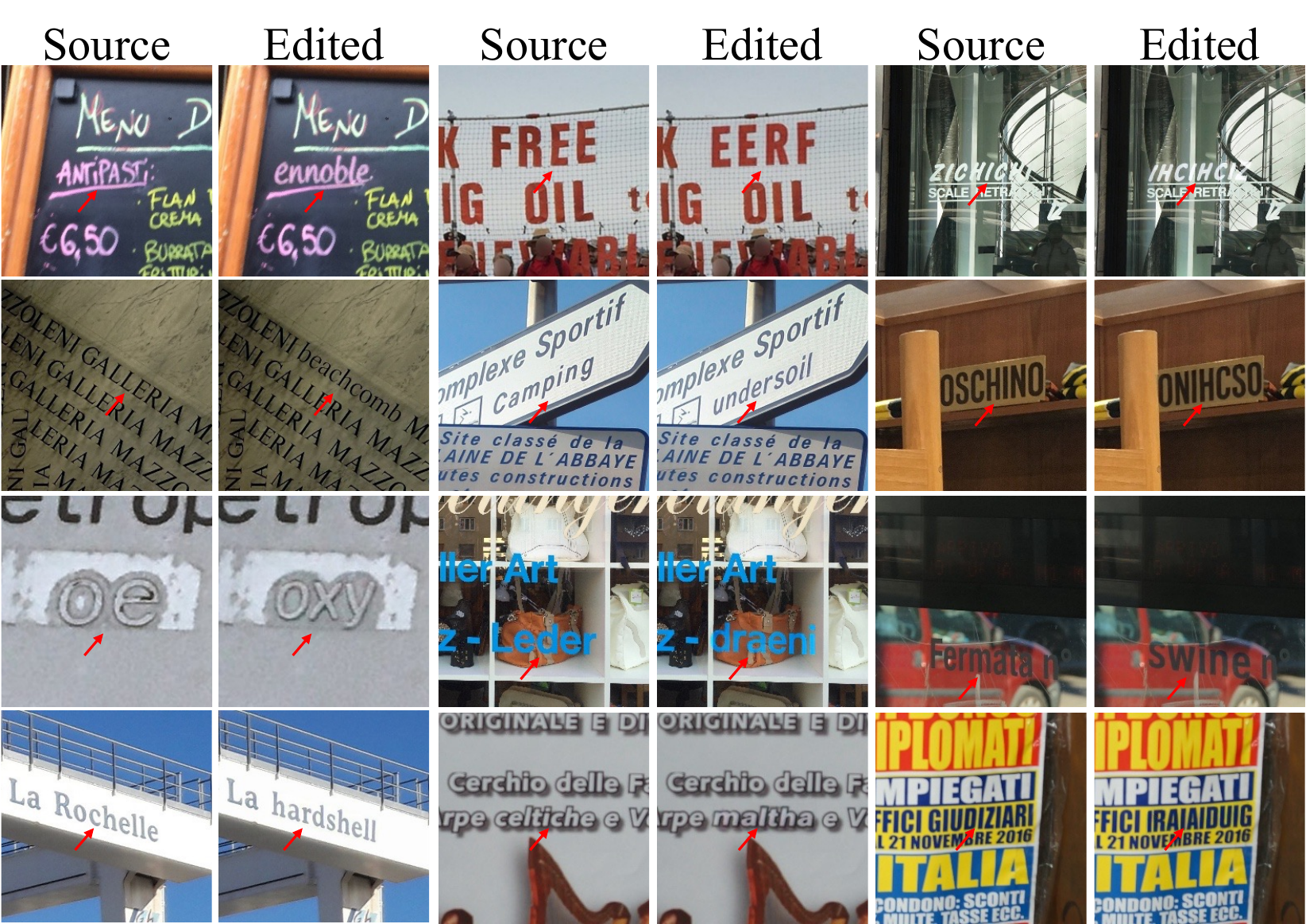}
    \caption{ 
    Examples of scene text editing results of our GLASTE.
    }
    \label{fig:examples}
    \vspace{-1.2em}
\end{figure}

\begin{figure}[t]
    \centering
\includegraphics[width=0.45\textwidth]{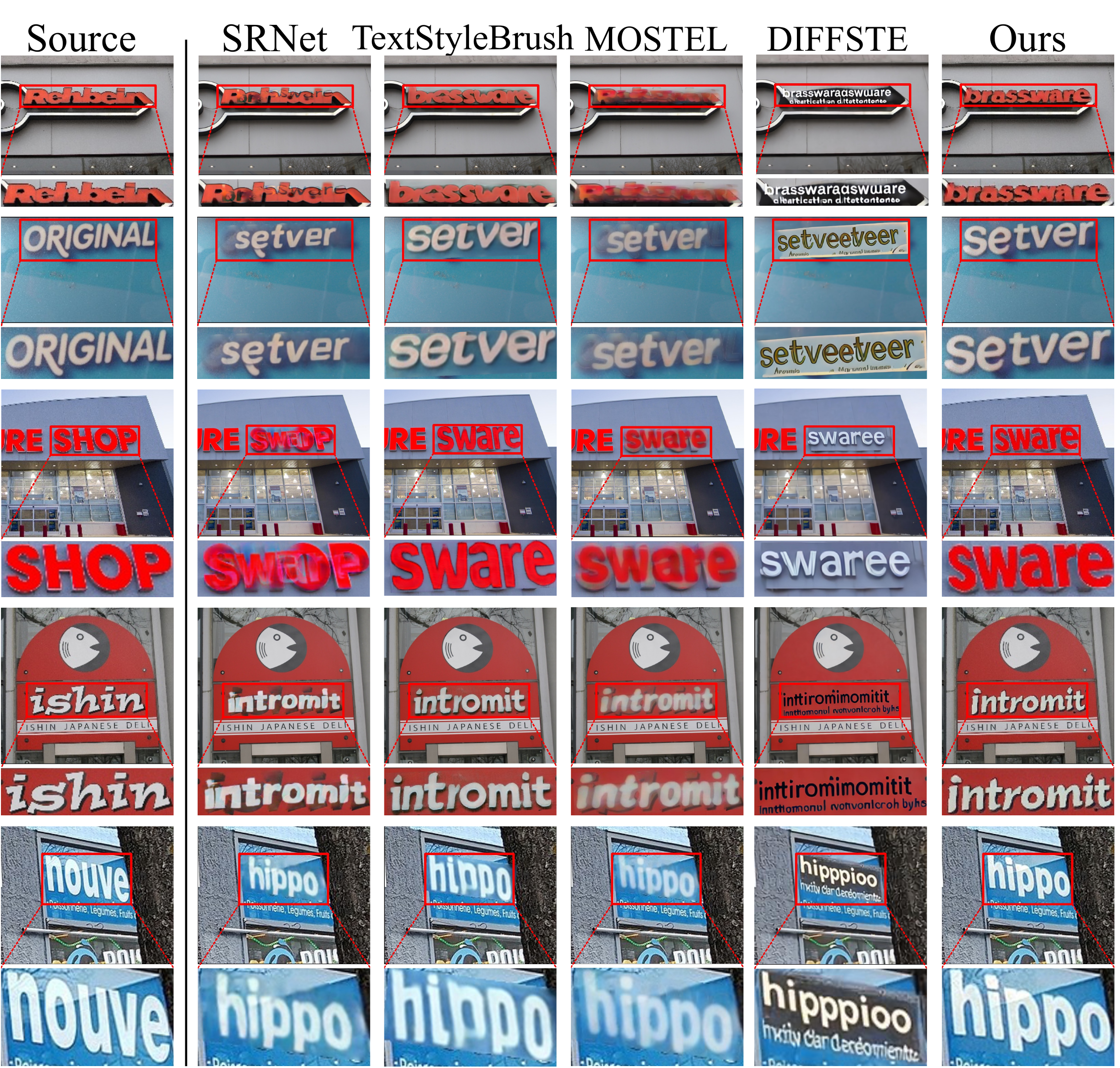}
    \caption{ 
     Comparison of previous methods and our GLASTE.
    }
    \label{fig:compare}
    \vspace{-1.2em}
\end{figure}

\section{Experiments}
\subsection{Experiment Setup}

\textbf{Datasets.} Our dataset includes a real-world dataset and a synthetic dataset. 
The real-world data is compiled from open-source text detection and recognition datasets, including ICDAR2015, MLT2017, MLT2019, ICDAR2019, SROIE, and ICDAR2017rctw. 
The training set has 120,450 images and the test set has 2,000 images. 
The synthetic data is generated with reference to SynthText \cite{gupta2016synthetic}, using our own collected English lexicons, over 500 fonts, and background images without text.
The training set has 200,000 images and the test set has 1000 images.
For real-world data, paired data is unavailable, so the text content of $I_{Ls}$, $T_c$, and $G_{Lc}$ are identical. 
However, for synthetic data, $T_c$ and $G_{Lc}$ contain new target text that is different from $I_{Ls}$, which is generated by pairing text samples and applying consistent parameters like font, color, and size to text rendered over background images.

\textbf{Implementation Details.}
During training, we mix the real data and synthetic data at a 1:1 ratio, with a batch size of 12, training for 500,000 iterations. 
$H_c$ is scaled to 64, and $W_c$ is scaled proportionally to the average width of the batch, and we set $H_L=W_L=256$. 
The Adam optimizer is used with default parameters and a learning rate of $10^{-4}$. 
To balance the weights of losses, we empirically set them to: $\alpha=1, \lambda_{1}=10, \lambda_{2}=1, \lambda_{3} = 0.1$, $\beta=10$.

\textbf{Evaluation Metrics.} We adopt several commonly used image generation evaluation metrics on target text patches. These include MSE, which measures the $L_{2}$ distance; PSNR, which computes the peak signal-to-noise ratio; SSIM, which calculates the structural similarity; Fréchet Inception Distance (FID) \cite{heusel2017gans}; Learned Perceptual Image Patch Similarity (LPIPS) \cite{zhang2018unreasonable}; and text recognition accuracy (Acc) and Character Error Rate (CER), where we use the text recognition model CRNN \cite{shi2016end} for evaluation.

\begin{figure}[t]
    \centering
    \includegraphics[width=0.45\textwidth]{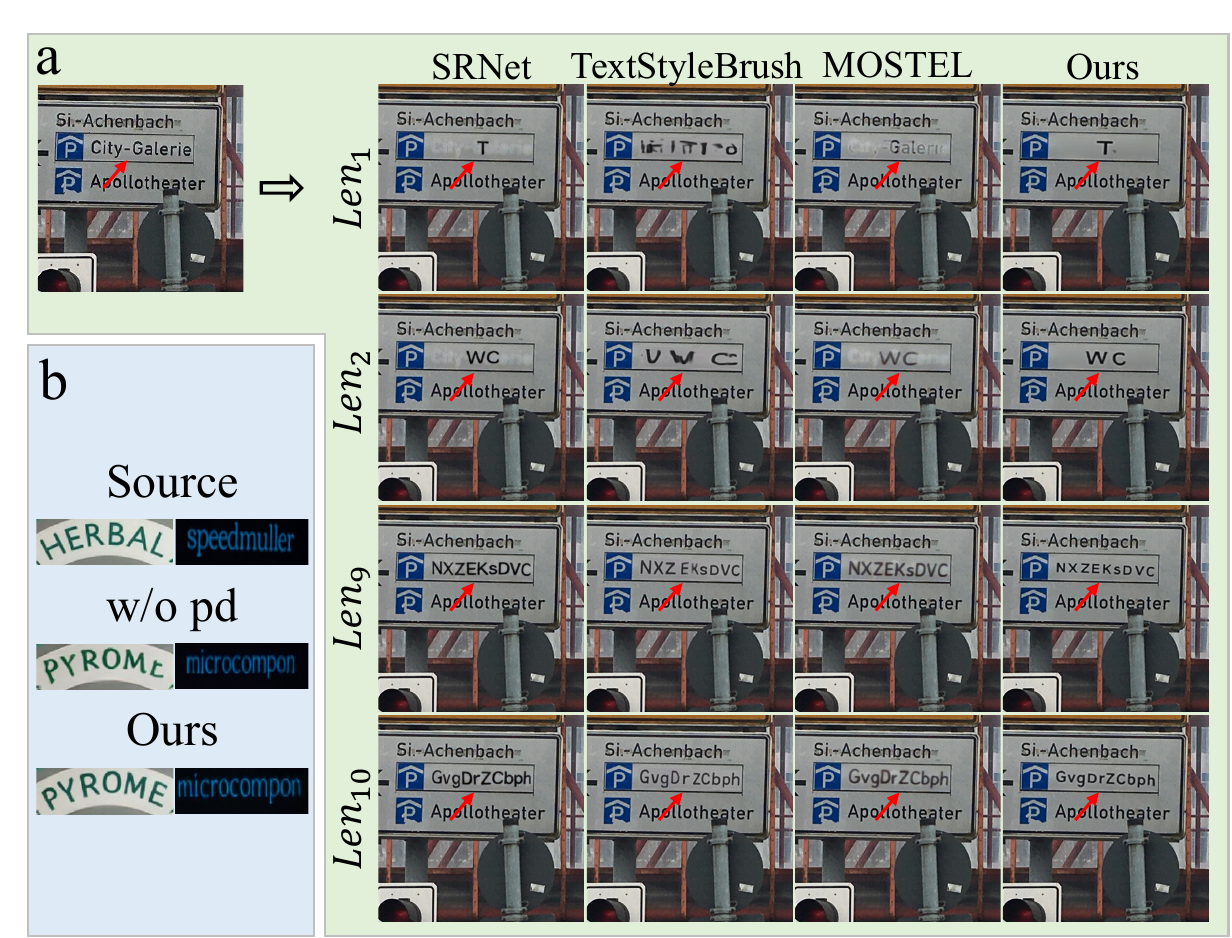}
    \caption{
     a) Editing from fixed source text to variable length target texts.
     b) Use synthetic paired data to alleviate overfitting.
     ``w/o pd'' means ``w/o paired data''.
    }
    \label{fig:length}
    \vspace{-1.2em}
\end{figure}

\begin{table*}[t]
\caption{Quantitative comparison between STE methods on real-world dataset.}
\label{merged_comparison}
\centering
\scalebox{0.72}{
\begin{tabular}{lcccccc|cc|cccc|cc} 
\toprule
\multirow{3}{*}{Method} & \multicolumn{6}{c}{Reconstruct \( I_{Ls} \)} & \multicolumn{2}{c}{Generate target texts} & \multicolumn{6}{c}{Generate target texts of different lengths} \\ 
\cmidrule(lr){2-7} \cmidrule(lr){8-9} \cmidrule(lr){10-15}
  & \multirow{2}{*}{MSE $\downarrow$} & \multirow{2}{*}{PSNR $\uparrow$} & \multirow{2}{*}{SSIM $\uparrow$} & \multirow{2}{*}{LPIPS $\downarrow$} & \multirow{2}{*}{FID $\downarrow$} & \multirow{2}{*}{Acc (\%) $\uparrow$} & \multirow{2}{*}{FID $\downarrow$} & \multirow{2}{*}{Acc (\%) $\uparrow$} & \multicolumn{4}{c}{Acc (\%) $\uparrow$} & \multicolumn{2}{c}{CER $\downarrow$}    \\
 &  &  &  &  &  &  &  & & $Len_1$   &  $Len_2$  & $Len_9$ & $Len_{10}$ & $Len_9$ & $Len_{10}$  \\ 
\midrule
SRNet~\cite{wu2019editing} & $0.336 {\pm \text{\tiny 0.006}}$ & $16.9 {\pm \text{\tiny 0.287}}$ & $0.505 {\pm \text{\tiny 0.009}}$ & $0.258 {\pm \text{\tiny 0.004}}$ & $26.9 {\pm \text{\tiny 0.457}}$ & $88.2 {\pm \text{\tiny 1.5}}$ & $46.2 {\pm \text{\tiny 0.785}}$ & $74.6 {\pm \text{\tiny 1.3}}$ & $33.5 {\pm \text{\tiny 0.6}}$ & $41.2 {\pm \text{\tiny 0.7}}$ & $35.3 {\pm \text{\tiny 0.6}}$ & $32.4 {\pm \text{\tiny 0.6}}$ & $0.314 {\pm \text{\tiny 0.005}}$ & $0.346 {\pm \text{\tiny 0.006}}$ \\ 
TextStyleBrush~\cite{krishnan2023textstylebrush} & $0.227 {\pm \text{\tiny 0.003}}$ & $17.8 {\pm \text{\tiny 0.263}}$ & $0.563 {\pm \text{\tiny 0.008}}$ & $0.250 {\pm \text{\tiny 0.004}}$ & $27.3 {\pm \text{\tiny 0.403}}$ & $81.8 {\pm \text{\tiny 1.2}}$ & $28.6 {\pm \text{\tiny 0.423}}$ & $58.7 {\pm \text{\tiny 0.9}}$ & $1.10 {\pm \text{\tiny 0.0}}$ & $13.6 {\pm \text{\tiny 0.2}}$ & $17.6 {\pm \text{\tiny 0.3}}$ & $12.2 {\pm \text{\tiny 0.2}}$ & $0.391 {\pm \text{\tiny 0.006}}$ & $0.444 {\pm \text{\tiny 0.007}}$ \\ 
MOSTEL~\cite{qu2023exploring} & $0.280 {\pm \text{\tiny 0.004}}$ & $17.9 {\pm \text{\tiny 0.237}}$ & $0.491 {\pm \text{\tiny 0.006}}$ & $0.274 {\pm \text{\tiny 0.004}}$ & $44.6 {\pm \text{\tiny 0.590}}$ & $82.4 {\pm \text{\tiny 1.1}}$ & $57.1 {\pm \text{\tiny 0.755}}$ & $64.1 {\pm \text{\tiny 0.8}}$ & $2.70 {\pm \text{\tiny 0.0}}$ & $20.2 {\pm \text{\tiny 0.3}}$ & $46.0 {\pm \text{\tiny 0.6}}$ & $34.1 {\pm \text{\tiny 0.5}}$ & $0.247 {\pm \text{\tiny 0.003}}$ & $0.32 {\pm \text{\tiny 0.004}}$ \\ 
DIFFSTE~\cite{ji2023improving} & $0.923 {\pm \text{\tiny 0.013}}$ & $11.1 {\pm \text{\tiny 0.156}}$ & $0.203 {\pm \text{\tiny 0.003}}$ & $0.476 {\pm \text{\tiny 0.007}}$ & $54.2 {\pm \text{\tiny 0.761}}$ & $39.8 {\pm \text{\tiny 0.6}}$ & $71.0 {\pm \text{\tiny 0.997}}$ & $7.90 {\pm \text{\tiny 0.1}}$ & \multicolumn{4}{c}{-} & \multicolumn{2}{c}{-} \\ 
\midrule
\textbf{Ours} & \textbf{0.108$ {\pm \text{\tiny 0.001}}$} & \textbf{22.4$ {\pm \text{\tiny 0.285}}$} & \textbf{0.721 ${\pm \text{\tiny 0.009}}$} & \textbf{0.129$ {\pm \text{\tiny 0.002}}$} & \textbf{12.0$ {\pm \text{\tiny 0.153}}$} & \textbf{96.3 ${\pm \text{\tiny 1.2}}$} & \textbf{24.4$ {\pm \text{\tiny 0.311}}$} & \textbf{83.7 ${\pm \text{\tiny 1.1}}$} & \textbf{84.3$ {\pm \text{\tiny 1.1}}$} & \textbf{88.2$ {\pm \text{\tiny 1.1}}$} & \textbf{65.7$ {\pm \text{\tiny 0.8}}$} & \textbf{61.7$ {\pm \text{\tiny 0.8}}$} & \textbf{0.073$ {\pm \text{\tiny 0.001}}$} & \textbf{0.086 ${\pm \text{\tiny 0.001}}$} \\ 
\bottomrule
\end{tabular}
}
\vspace{-1.2em}
\end{table*}

\subsection{Comparison with Previous Work}
We compare our model, GLASTE, with four other models: SRNet \cite{wu2019editing}, TextStyleBrush \cite{krishnan2023textstylebrush}, MOSTEL \cite{qu2023exploring}, and DIFFSTE \cite{ji2023improving}. Among them, the first three are GAN-based and focus on generating local text patches, while DIFFSTE, a diffusion-based model, generates text patches from a global perspective. 
Figure \ref{fig:examples} shows scene text editing results of GLASTE on real-world datasets. Visual comparisons are presented in Figure \ref{fig:compare}, and Table \ref{merged_comparison} summarizes the quantitative metrics derived from three repeated trials. 

Our method achieves superior performance across all metrics and visual effects, both in reconstructing \( I_{Ls} \) and generating target texts.
First, GLASTE significantly reduces shadow artifacts from the original text. As shown in the first three examples in Figure \ref{fig:compare}, SRNet produces the most severe shadow artifacts, followed by MOSTEL. By leveraging a global inpainting module, GLASTE recovers pixels inside the text patch, ensuring a cleaner background texture and minimizing interference from the original text.
Second, GLASTE generates higher-quality text patches that closely match the source style, yielding globally consistent visual effects and the best quantitative results, including MSE (0.108), PSNR (22.4), SSIM (0.721), LPIPS (0.129), and FID (12.0/24.4). In contrast, GAN-based methods, such as SRNet, TextStyleBrush, and MOSTEL, focus only on local patch generation, resulting in lower-quality outputs. Among them, TextStyleBrush performs best in generating target text patches, with an FID of 28.6. However, noticeable discrepancies with the original style remain, such as missing 3D effects (examples 1 and 3 in Figure \ref{fig:compare}) and blurriness (last example in Figure \ref{fig:compare}).
Although DIFFSTE also edits text patches globally, it often produces text styles inconsistent with the original image. 
Finally, GLASTE achieves the highest text recognition accuracy (96.3/83.7). Among the other methods, DIFFSTE performs worst (39.8/7.90) as it is more suited for generating semantically meaningful words. Random character combinations often lead to repetition or omissions, resulting in lower accuracy.

\subsection{Robustness to Variable Length}

To verify whether the model can adapt to cases where the text length differs significantly before and after editing, we randomly select target texts of lengths 1, 2, 9, and 10 for each sample in the real data test set.
Due to the poor recognition accuracy of DIFFSTE in generating text images with random character combinations and its slow sampling speed in the tests above, we only compare with SRNet \cite{wu2019editing}, TextStyleBrush \cite{krishnan2023textstylebrush} and MOSTEL \cite{qu2023exploring}.
The visualization results are shown in Figure \ref{fig:length} (a), and the quantitative results are shown in Table \ref{merged_comparison}.

Our GLASTE method demonstrates significant advantages over other approaches. Competing methods struggle to handle text length transformations effectively. For short target texts, SRNet performs best, achieving accuracies of 33.5 and 41.2 for lengths 1 and 2, respectively. For long target texts, MOSTEL performs best, with accuracies of 46.0 and 34.1 for lengths 9 and 10. However, all these results fall short of our method. GLASTE achieves very high recognition accuracies of 84.3 and 88.2 for target text lengths of 1 and 2. Although the recognition accuracy for target text lengths of 9 and 10 is relatively lower (65.7, 61.7), the CER remains very low at 0.073 and 0.086, demonstrating our model's robustness.

\subsection{Ablation Study}

In this section, we evaluate the impact of key model settings on synthetic paired data. Quantitative metrics are presented in Table \ref{Ablation-table_gen-target}, focusing on the quality of the target text region.

\textbf{Global-Local Combination Effectiveness.}  
As shown in Figure \ref{fig:architecture}, all inputs and outputs, except for the local text patches \(T_c\) and \(G_f\), are large scene images. When \(T_c\) is affine-transformed into a larger image (``w/ transform \(T_c\)''), the model adopts a more global perspective but suffers an accuracy drop of 9.9 (from 92.3 to 82.4). This decline occurs because most pixels in \(T_c\) and \(G_f\) are zeros, leading to inefficient parameter utilization. Therefore, we retain \(T_c\) and \(G_f\) as local patches while employing a global design for other modules.

\textbf{Effectiveness of AdaIN.}  
Our approach leverages AdaIN to inject style into content text images, which is both simple and effective. Although cross-attention can also introduce conditions such as text and images~\cite{rombach2022high}, replacing AdaIN with cross-attention (``w/ cross attention'') results in a performance drop across all metrics and fails to generate readable text (Accuracy: only 4.9).

\textbf{Enhancing Text Image Features.}  
We introduce skip connections to enhance text feature maps, leading to better recognition performance. Removing this structure (``w/o skip connection'') causes a significant accuracy drop of 26.7 (from 92.3 to 65.6), underscoring its importance.

\textbf{Relationship Between Global and Local Losses.}  
Our global loss function is crucial for maintaining quality. Omitting local losses (``w/o local loss,'' i.e., setting $\alpha=\beta=\lambda_3=0$) degrades the generated local text patches. LPIPS increases by 0.045 (from 0.292 to 0.337), FID worsens by 8.6 (from 39.8 to 48.4), and accuracy drops by 27.9 (from 92.3 to 64.4). Combining global and local losses yields improved results.

\textbf{Resistance to Overfitting.}  
Real-world data is unpaired, and using $ I_{Ls} $ as supervision with only real data often leads to slight overfitting, causing the generated output to embed excessive information from $ I_{Ls} $. As shown in Figure~\ref{fig:length} (b), where ``w/o paired data'' renders the ``E'' in ``PYROME'' consistent with the ``L'' in the Source, and ``microcompon'' matches the rendering width of the source image.

\textbf{Recognition Model Contribution.}  
Incorporating CTC loss enhances text recognition accuracy. While removing the recognizer (``w/o R'') improves other metrics, it reduces recognition accuracy by 3.0 (from 92.3 to 89.3).

\begin{table}[t]
    \caption{Ablation study on synthetic dataset.}
    \label{Ablation-table_gen-target}
    \centering
    \scalebox{0.8}{
    \begin{tabular}{lcccccc} 
    \toprule
      Method  & MSE $\downarrow$ & PSNR $\uparrow$ & SSIM $\uparrow$ &  LPIPS  $\downarrow$   &  FID $\downarrow$ &  Acc (\%) $\uparrow$ \\ 
      \midrule
      w/ transform $T_c$  & 0.323 & 16.2 & 0.398 & 0.301  & 41.5 & 82.4  \\
      w/ cross attention &  0.360 & 15.5 & 0.357 & 0.462  & 88.7 & 4.90  \\
      w/o skip connection & 0.315 & 16.2 & 0.404 & 0.298  & 40.5 & 65.6 \\
      w/o local loss & \textbf{0.295} & 16.4 & 0.402 & 0.337  & 48.4 & 64.4 \\
      w/o paired data & 0.349 & 15.8 & 0.387 & 0.305  & 42.7 & 90.3  \\ 
      w/o R & 0.301 & \textbf{16.5} & \textbf{0.426} & \textbf{0.276}  & \textbf{38.3} & 89.3 \\
      \midrule
      Ours & 0.312 & 16.3 & 0.406 & 0.292 & 39.8 & \textbf{92.3} \\ 
      \bottomrule
    \end{tabular}
    }
    \vspace{-1.2em}
\end{table}

\section{Conclusion}
This paper proposes a Global-Local Aware Scene Text Editing (GLASTE) model to tackle both the inconsistency and length-insensitivity challenges. 
Firstly, given that the text patch occupies a relatively small portion of the scene image, we wisely designed the model architecture and loss functions to ensure clear, reasonable, and recognizable text structure in the edited patch while maintaining harmony between local and global areas.
Secondly, our model can flexibly adapt to varying target text lengths during inference, accounting for both lengthening and shortening editing scenarios.
Through extensive experiments on real-world datasets, we have validated that our model outperforms previous methods in terms of quantitative performance metrics.

The limitations of our work lie in not utilizing diffusion models as the generative framework. 
However, compared to GAN-based methods, diffusion model-based methods require substantial computational resources and time for both training and inference. 
Moreover, we found that diffusion-based methods struggle with generating text from random character combinations (such as DIFFSTE), often repeating or omitting characters.
This indicates the need for further refinement at the character level for such methods.

\section{ACKNOWLEDGEMENT}
This work was supported by the National Key Research and Development Program of China (Grant No. 2020AAA0108003) and National Natural Science Foundation of China (Grant No. 62277011).

\bibliographystyle{IEEEbib}
\bibliography{icme2025references}

\end{document}